\documentclass{article}

\usepackage{microtype}
\usepackage{graphicx}
\usepackage{subfigure}
\usepackage{booktabs} 
\usepackage{floatrow}
\usepackage{natbib}
\usepackage{hyperref}

\usepackage[accepted]{icml2024}

\usepackage{amsmath}
\usepackage{amssymb}
\usepackage{mathtools}
\usepackage{amsthm}

\theoremstyle{plain}

\theoremstyle{definition}

\theoremstyle{remark}

\usepackage[textsize=tiny]{todonotes}

\icmltitlerunning{Attention Track: Prototype-based XAI and Opportunities in Geosciences}

\begin{document}

\twocolumn[
\icmltitle{Prototype-Based Methods in Explainable AI and Emerging Opportunities in the Geosciences}




\begin{icmlauthorlist}
\icmlauthor{Anushka Narayanan}{dsi,deeps}
\icmlauthor{Karianne J.\ Bergen}{dsi,deeps,cs}
\end{icmlauthorlist}

\icmlaffiliation{dsi}{Data Science Institute, Brown University, Providence, RI, USA.}
\icmlaffiliation{deeps}{Department of Earth, Environmental and Planetary Sciences, Brown University, Providence, RI, USA.}
\icmlaffiliation{cs}{Department of Computer Science, Brown University, Providence, RI, USA.}

\icmlcorrespondingauthor{Anushka Narayanan}{anushka\_narayanan@brown.edu}

\icmlkeywords{explainable AI, geosciences, case-based reasoning, prototype-based methods}

\vskip 0.3in
]



\printAffiliationsAndNotice{} 

\begin{abstract}

Prototype-based methods are intrinsically interpretable XAI methods that produce predictions and explanations by comparing input data with a set of learned prototypical examples that are representative of the training data. In this work, we discuss a series of developments in the field of prototype-based XAI that show potential for scientific learning tasks, with a focus on the geosciences. We organize the prototype-based XAI literature into three themes: the development and visualization of prototypes, types of prototypes, and the use of prototypes in various learning tasks. We discuss how the authors use prototype-based methods, their novel contributions, and any limitations or challenges that may arise when adapting these methods for geoscientific learning tasks. We highlight differences between geoscientific data sets and the standard benchmarks used to develop XAI methods, and discuss how specific geoscientific applications may benefit from using or modifying existing prototype-based XAI techniques. 
\end{abstract}

\section{Introduction}

Machine learning (ML) and deep learning (DL) are powerful tools for modeling, classification, and prediction. In recent years, ML/DL has been widely adopted by scientists because these tools outperform the best domain-specific, non-ML methods with low computational costs and improved scalability for many tasks in scientific data analysis \citep{wang2023scientific}. Weather and climate science  is an example of a scientific domain that has recently seen a rapid adoption of ML and DL methods \cite{rolnick2022tackling,reichstein2019deep}. ML/DL has driven impressive advances in weather forecasting models \citep{pathak2022fourcastnet, lam2023learning,bi2023accurate}, climate emulators \citep{beucler2021enforcing}, extreme event prediction \citep{griffin2022predicting}, spatio-temporal forecasting \citep{nguyen2023climax}, and remote sensing image analysis \citep{jakubik2023foundation}.  However, a common argument against the unrestrained use of ML in scientific applications is that most ML/DL methods operate as ``black-box" models that lack ``interpretability" \citep{rudin2019stop, de2023machine}. In scientific research, specifically in geoscientific disciplines, interpretable ML/DL models and explainable AI (XAI) techniques are critical to verifying the model has learned the correct underlying  physical principles and patterns governing the data \citep{yang2024interpretable}. This is especially important to build trust in models deployed in operational settings, such as early warning systems for extreme weather \citep{kuglitsch2022artificial}. Interpretability is also key when ML is used with the aim of discovering novel scientific insights or patterns in data to advance scientific research \citep{mcgovern2019making}. Accounting for interpretability allows researchers to go beyond predictive accuracy and showcase the equally important insight of how or why the model makes specific predictions. \\
 
In scientific research, including in the fields of climate and weather, the most commonly used interpretability methods are \textit{post hoc} techniques \citep{toms2020physically, ham2019deep}. Post hoc XAI techniques such as SHAP \citep{lundberg2017unified}, LIME \citep{ribeiro2016should}, or LRP \citep{bach2015pixel} allow scientists to investigate feature patterns, quantify predictor importance, or generate saliency maps. However post hoc techniques have certain known limitations and drawbacks \citep{zhou2021evaluating,neely2021order,adebayo2018sanity}. \citet{lipton2018mythos} highlights the difficulty of interpreting predicitions when different XAI methods yield conflicting explanations. In an example from the geosciences, \citet{mamalakis2022investigating} compare a series of post hoc XAI methods to explain the decisions of a convolutional neural network (CNN) for a climate-related prediction task. The authors show that each post hoc method included in the analysis resulted in inconsistent and complex model explanations. \citet{mcgovern2019making} caution against using post hoc XAI methods to derive scientific conclusions as rigorous hypothesis testing is required to ensure the viability of the insights. 

In contrast, \textit{inherently interpretable} models can alleviate some of the concerns associated with post hoc methods. An inherently interpretable model aims to explain its reasoning as it makes its prediction and does not require additional techniques to interpret the prediction after-the-fact \citep{rudin2019stop}. Linear regression (interpretable feature weights) and decision trees (binary decision rules) \citep{gilpin2018explaining} are two examples of ML methods that can be classified as inherently interpretable. However, more complex models involving DL architectures, such as CNNs, ANNs and LSTMs, generally do not share a similar inherent interpretability. \\

\textit{Case-based} reasoning can be used as a way to integrate interpretability into complex model architectures. Case-based reasoning uses the comparison between an input and a particular instance to explain a prediction \citep{keane2019case}, such as instances from the training data or counterfactual examples. A popular example of case-based reasoning is the use of learned \textit{prototypes} embedded in the model architecture to reason out the model's predictions. Generally, the model identifies \textit{prototypical representations} of the data during training and uses features representing the similarity between an input and the learned prototypes to make a prediction \citep{li2018deep}. In this approach, the user can interpret the model's prediction by inspecting the most similar prototype(s), which represent typical features or patterns in the data. This approach to generating explanations aims to mimic the human reasoning process; for example, one might identify a bird's species by comparing key traits: its beak, wing, and feet, to an example of a ‘prototypical’ beak, wing or feet for a sparrow (and other candidate species) \citep{chen2019looks}.  Prototype-based explanation techniques offer an alternative to common post hoc XAI techniques, with the added benefit  that the reasoning provided by prototype-based explanations is inherently built into the model decision making process.  \\

Prototype-based XAI techniques are an underutilized approach that can provide inherently interpretable ML alternatives for the scientific research community already using ML. This review paper is organized as follows. In Section \ref{sec:prototypes} we present a brief overview of prototypes. In Section \ref{sec:casestudies}, we review methods for prototype-based XAI methods and categorize the literature into (1) studies that focus on the development and visualization of prototypes, (2) studies that derive different types of prototypes, and (3) studies that use prototypes for different types of learning tasks. In Section \ref{sec:dicussion}, we present a perspective on how scientists can leverage and extend these methods in their research. Specifically, we highlight research tasks in the domain of climate and weather that may benefit from leveraging case-based XAI techniques where explanations are derived from computing a similarity metric to a case or previous instance. Many climate and weather research studies currently use post hoc XAI techniques to interrogate ML models \citep{yang2024interpretable}. We argue that the application and development for prototype-based inherently interpretable XAI approaches for geoscientific data is a promising avenue for future research toward advancing scientific discovery in weather and climate science.

\section{Prototype-based XAI Models}\label{sec:prototypes}

Before examining specific prototype XAI methods, we describe the general architecture of prototype-based neural networks. A prototype-based model architecture consists of standard neural network layers (e.g., convolution, recurrent, dense) that learn representative features of the dataset, a \textit{prototype layer} that generates latent prototypes associated with each class in training and computes a similarity metric between the input and the prototypes, and a fully connected component that converts the similarity scores to the prototypes to a final output specific to the learning task (e.g., classification, regression, prediction) \citep{li2018deep}. The model also needs a component that allows visualization of the latent prototypes; this latent space visualization component is necessary because prototypes are representations in the latent space, which can be difficult to interpret directly, so we project these representations to visualize them in the same space as the data. \\

In addition to the standard trainable parameters a neural network learns, the prototypes themselves are represented by learnable parameters that are iteratively updated according to a specified optimization procedure \citep{chen2019looks}, which typically includes additional terms in the loss function. Generally, the loss function is designed to ensure explanations are meaningful: the prototypes should be human-interpretable, represent diverse representations of a target class \citep{chen2019looks}, and be similar to instances in the training dataset \citep{li2018deep}. This is essential to the case-based reasoning process, where a sample’s prediction derives its explanations from its similarity to previously seen instances (i.e., in the case of ML, from the training dataset). When a prototype network is deployed on test samples, the model's final prediction is derived from the prototype with the highest similarity (i.e., predicting the target label of that specific prototype with the highest similarity), thus providing an explanation for its prediction.

\section{Case Studies} \label{sec:casestudies}
The studies discussed below are organized into three categories. Section \ref{sec:casestudies-dev} traces the development of prototype-based XAI networks and how we visualize the prototypes to make them human-readable. Section \ref{sec:casestudies-types} focuses on case studies that develop novel additions to the prototype network to generate different prototype-based explanations. Section \ref{sec:casestudies-tasks} presents studies focused on developing novel ways to use the similarity metrics and discusses how they can inform the learning task and the relevant explanation. In each of the case studies, we will discuss the study’s novel contribution to the field, their specific methodology and application, and any potential limitations or challenges that can arise when applying these tools for geoscience-related research. 

\subsection{Development and Visualization of Prototypes}\label{sec:casestudies-dev}

\subsubsection{Image-Sized Prototypes}
\begin{figure}[htb!]
    \centering
    \includegraphics[width=\columnwidth]{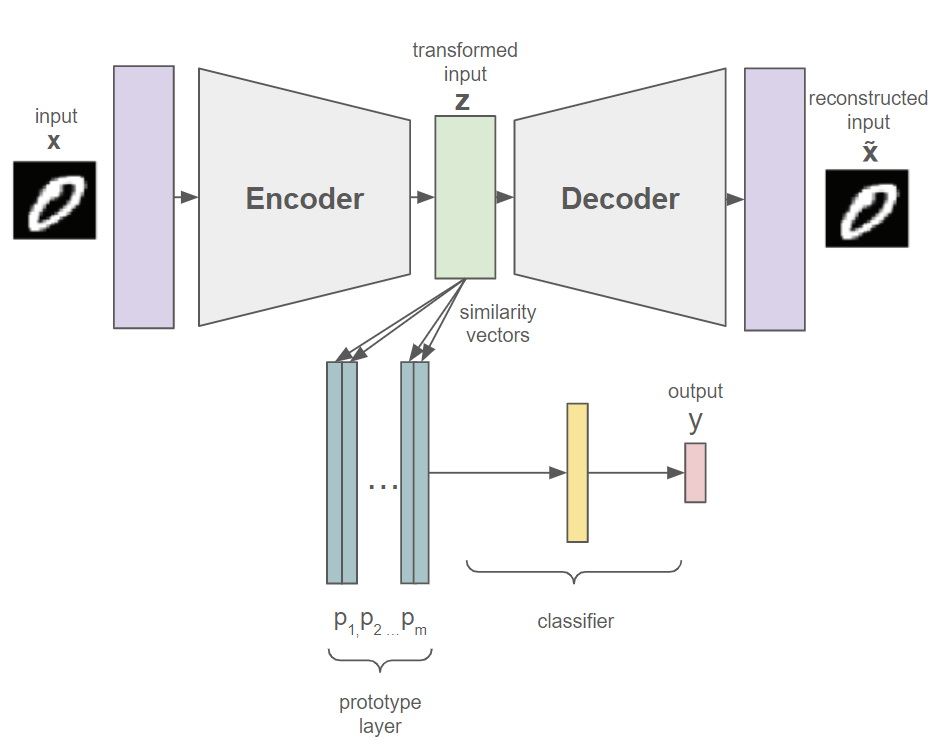}
    \caption{Diagram of the prototype-based XAI model \cite{li2018deep} with encoder-decoder architecture and classifier for an MNIST example.}
    \label{fig:vanilla-protonet}
\end{figure}

\citet{li2018deep} develop a prototype-based image classifier, as defined in Section \ref{sec:prototypes}, for XAI with case-based reasoning as a primary objective. The authors train their model to classify handwritten digits in the benchmark MNIST image dataset using an encoder-decoder architecture with an additional classifier network (see Figure \ref{fig:vanilla-protonet}).  The encoder learns useful and relevant features in the image for both the classification task and image reconstruction. The prototype layer learns prototypical vectors in the latent space from the encoder. These prototypes are updated via the loss constraints described in Section \ref{sec:prototypes}. The prototype layer outputs a similarity metric between the input image and the set of learned prototypes. The similarity scores are used in final softmax classifier. The prototypes are visualized with the decoder network, which maps the latent prototype vectors back to the original input dimension such that the decoded prototypes are of the original input image size. The authors successfully demonstrate their prototype-classifer network achieve competitive accuracy on benchmark datasets such as MNIST in comparison with a standard CNN without the use of prototypes. The authors demonstrate a limitation of this work: the decoded prototypes are not guaranteed to correspond to realistic or plausible instances. As a result, this approach may produce indecipherable or incomprehensible prototypes images, which may not provide a human-understandable explanation. 

\subsubsection{ProtoPNet: Patch-based Prototypes}
\citet{chen2019looks} follow up on the previous study and introduce ProtoPNet, a prototypical-part-based network that is presently widely cited as the standard prototype-based network in case-based reasoning XAI. The key innovation of \citet{chen2019looks} is an architecture that allows the prototypes to represent ``parts" (or patches) of the input images, rather than the whole image as in \cite{li2018deep}. The latent space consists of the output of the convolution layers broken up into a grid of patches. The loss function ensures the training images contain latent patches similar to prototypes of the correct class and ensures the prototype patches are dissimilar to prototypes of other classes. In ProtoPNet, the authors constrain the prototype patches to be $1\times{}1\times{}D$ where $D$ is the number of channels from the convolution output. This allows for prototypes to represent typical attributes of local features rather than the entire image itself. Another contribution of the work is the use of a projection technique that requires the latent prototype patch to be identical to a training sample rather than just maximizing the similarity to the training sample. In doing so, the authors remove the necessity of the decoder network and rather project the latent prototypical patch onto a training sample. Since each prototype is constrained to correspond to part of the feature map from a training sample, we can trace the latent prototype back to an actual patch in the training images, which is easily visualizable and human-interpretable. The authors demonstrate their prototype network model on a bird image classification task, as the example in the introduction, where prototypes of each class (i.e., sparrow) may represent a red beak, a black eye, or other distinct, localized visual characteristics. However, the authors state in the cases where test images contain similarities to prototypical parts of multiple classes, those prototypical parts tend to correspond to patterns in the background of the image. This mandates users to employ pruning approaches to ensure prototypes consist of patterns from the object of interest and not necessarily the background.

\subsubsection{Spatially Deformable Prototypes}
\begin{figure}[htb!]
    \centering
    \includegraphics[scale=0.6]{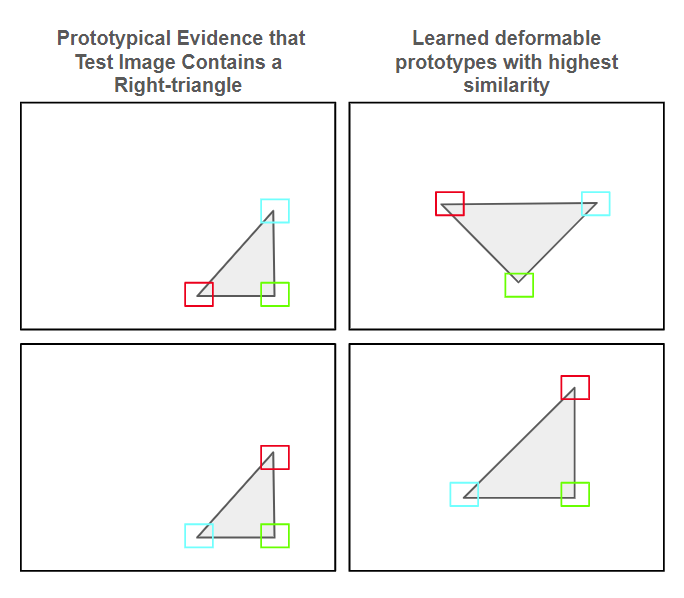}
    \caption{Notional diagram of deformable prototypes \cite{donnelly2022deformable} where the prototypical patches (bounding boxes) \cite{chen2019looks} vary in their spatial organization for one input image (see Fig. 5 in original paper \cite{donnelly2022deformable}).}
    \label{fig:deformable}
\end{figure}

\citet{donnelly2022deformable} extend ProtoPNet to produce a deformable prototype-based architecture that consists of similarly derived patch-based prototypes from the latent space as mentioned above in \citet{chen2019looks} that can be organized in a spatially flexible manner. The authors demonstrate the method on the benchmark bird classification dataset used in \citet{chen2019looks}. In this work, the authors allow a prototype to consist of smaller rectangular prototypical patches that can change their relative spatial positioning for an input image (see Figure \ref{fig:deformable}). The smaller rectangular prototypical patches are additionally constrained to be orthogonal to each other along with the orthogonal constraints among the larger prototypes themselves.  The spatially flexible prototypes allow for similarities between features in the input image and the prototypical patches to vary in their orientation, allowing for flexibility and learning of distorted or obscured images. However, the additional constraints may increase training complexity, and the resulting explanations have added complexity making them potentially less human-interpretable.

\subsection{Types of Prototypes}\label{sec:casestudies-types}

\subsubsection{ST-ProtoPNet: Trivial and Support Prototypes}

\begin{figure}[htb!]
    \centering
    \includegraphics[scale=0.6]{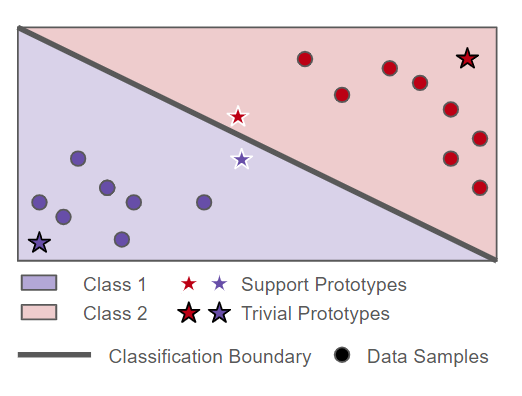}
    \caption{Notional diagram of support prototypes found closer to the classification boundary and trivial prototypes found far from each other (see Fig. 1 in original paper \cite{wang2023learning}).}
    \label{fig:supportproto}
\end{figure}

\citet{wang2023learning} present a technique where in addition to finding prototypes that are most representative of the classes in a dataset, termed \textit{trivial prototypes}, they allow prototypes to represent instances close to the classification boundary between the target classes. These prototypes are analogous to the support vectors in SVMs and thus termed \textit{support prototypes}. The trivial and support prototypes are notionally represented in Figure \ref{fig:supportproto} for binary classification, but this approach can be extended for multi-classification. The support prototypes of different classes are constrained to be close to the decision boundary between classes, in contrast with maximizing the distance between trivial prototypes of different classes. This removes the assumption that expects a test image to contain standard trivial prototypical parts of a class. Rather, samples that contain parts resembling support prototypes may contain subtle salient, distinguishing characteristics of the sample's target class. In an object classification example shown by the authors, trivial prototypes tend to focus on obvious distinct characteristics (i.e if the background is different between classes) compared to support prototypes that focused on salient characteristics of the foreground (i.e specific distinguishing features of the object). The support and trivial prototypes are alternatively optimized within each training epoch, and a weighted combination of similarity to the trivial prototypes and the support prototypes is used in the final classification. In the author's approach, the model is limited to learn an equal number of support and trivial prototypes and the same number of total prototypes for each class. This method may need to be modified for class-imbalanced data where a flexible number of support and trivial prototypes may be more useful. 

\subsubsection{ProtoSeNet: Sequential Prototypes}

\citet{ming2019interpretable} develop a network designed for analyzing sequential data in which prototypes represent a prototypical series of sequential events or ordered events in the training data. The similarities between sub-sequences in the input sample and the prototypical sub-sequences from previous cases are the explanations for the classification task. To develop prototypical sequences, the authors use a recurrent DL architecture that can capture important temporal and sequential information in the data. They ensure only critical sequences are generated as the final prototypes, avoiding duplicate prototypical sequences. The authors include a projection step that assigns the prototypical sub-sequence to the closest observed sub-sequence in the training data. In order to search through all possible sub-sequences of all lengths in each training sample, the authors employ a greedy search algorithm. However, the training complexity grows quadratically with increases in the observed sample's sequence length, which indicates this method may only be feasible for classifying sequences of shorter lengths. The authors apply their technique to text classification and protein sequence classification.  The authors also include a human-in-the-loop approach in which prototypes can be manually updated using user knowledge. In an experiment, they recruit non-ML experts to select prototypes that are most similar to a given target sequence (e.g., a sequence of text). The prototype options include both the highest similarity prototype from the model and other randomly chosen prototypes. If enough participants choose a prototype that the model did not choose, the prototype for that target class is manually updated. 

\subsubsection{Multi-variable Prototypes}

\citet{ghosal2021multi} present a network that aims to learn representative prototypes from multi-variable time series data. They develop single-variable prototypes by extracting prototypical features from each variable separately, similar to ProtoPNet in \citet{chen2019looks}. They also investigate the interactions between the single-variable prototypes to develop a multi-variable prototypical representation of the data. They first generate individual single-variable prototypical patterns, then they concatenate and extract relationships between the individual single-variable patterns to generate a comprehensive multi-variable prototype-based explanation. The authors test their approach on synthetically generated multi-variable time series data and on a benchmark Epilepsy time series dataset. They reveal that the prototypes capture patterns within each feature and relationships between multiple features. However a major limitation of this method is that the number of prototypes generated increases exponentially with the number of features in a dataset. For example, the authors use a synthetic dataset of 4 single-variable features (3 relevant to the target class, 1 random noise feature) in which the number of multi-variable prototypes is $4^3$, or 64, to account for all the combinations of the 3 relevant single-variables in the multi-variable prototype. This may be computationally infeasible to use for higher-dimensional data and can result in increasingly numerous explanations to navigate through for the user.   

\subsubsection{ProtoAD: Anomaly Detection in Time series Prototypes}

\citet{li2023prototypes} use a prototype-based approach to explain anomalies in time series data. The authors initially compute prototypical sequential data from training data consisting of ``normal" (non-anomalous) time series data using an LSTM-autoencoder in an unsupervised manner, using a reconstruction-based loss function. The authors detect anomalies in the time-series via anomaly scores. The anomaly score is computed by the model by comparing sequences in the time series to the prototypical sequences learned from the normal data. The authors do not prescribe a threshold for when an anomaly is detected; rather they evaluate performance metrics on the quality of the model-generated anomaly scores with real-valued anomaly scores. The authors apply their method to find anomalous behaviour in a synthetic test case of a sine wave function with random noise injections as well as various real-world benchmark datasets with competitive accuracies to non-prototype based anomaly detection methods. However, the authors' method requires the training data to be free of anomalies, which may prove difficult to guarantee for non-benchmark observed datasets without extensive domain-specific knowledge. 

\subsection{Prototype Reasoning Across Learning Tasks}\label{sec:casestudies-tasks}

\subsubsection{ProtoTree: Decision Trees for Learning Tasks}

\citet{nauta2021neural} present a combination of a prototype-based network and a decision-tree-based classification method, termed ProtoTree. The similarity scores to prototypes are passed as features into a decision tree that makes the final class prediction. For example, a simplified tree-based reasoning may look like a series of binary questions asking whether a bird image has high similarity to particular prototypes, which may be associated with different classes. In this example, the model first checks whether the input image contains the red beak prototype. If true, then it looks for a blue wing prototype. Through this reasoning process, the decision tree will output the classification. This allows for target classes to share similar prototypes (i.e both sparrow and robin have black eye color prototypes) and does not require the test image to contain all prototypes associated with a particular class. The tree-structure also allows for pruning of prototypes by removing leaves that have little discriminative power in the learning task. Pruning reduces the chance that multiple similar prototypes are included in the model. Compared to ProtoPNet, ProtoTree achieves similar accuracies with a significantly reduced number of prototypes, around 90\% of the prototypes needed for ProtoPNet. However, training must be performed in two stages: first to learn the optimal prototypes and then to learn the optimal splits in the decision tree. The authors point out that learning these parameters simultaneously resulted in an overly complex and inaccurate model. ProtoTree is able to generate local explanations by tracing a specific test instance's path through the decision tree. Global explanations of the model's reasoning for each target class can also be generated from the full decision tree. 

\subsubsection{NP-ProtoPNet: Learning Tasks with Negative Reasoning}

\citet{singh2021these} present NP-ProtoPNet, a method that closely resembles ProtoPNet \citep{chen2019looks} with the additional of a novel reasoning technique. ProtoPNet relies heavily on positive reasoning, where a network makes a classification based on how similar the patches in the input image are with a prototypical patch. In addition to positive reasoning, NP-ProtoPNet network uses negative reasoning: it can reject classes based on the absence of matching prototypical patterns between other classes and the input image. Similarity scores will be positive if the input contains to prototypical patterns of certain classes or negative if the input does not contain prototypical patterns of other classes. An equal combination of the positive and negative similarity scores is used to make final prediction. However, in this equal combination approach, models can tend to make predictions with using more negative reasoning, if the absolute value of the negative similarity scores are higher than the absolute value of the positive similarity scores (i.e the model is very confident the input does not look like Class A,B, or C but only mildly confident the input sample looks like Class D.) The authors demonstrate this method on the classification on X-ray images for the presence of Covid-19. 

\subsubsection{ProtoLNet: Learning Tasks with Location Scaling}

\begin{figure}[htb!]
    \centering
    \includegraphics[scale=0.5]{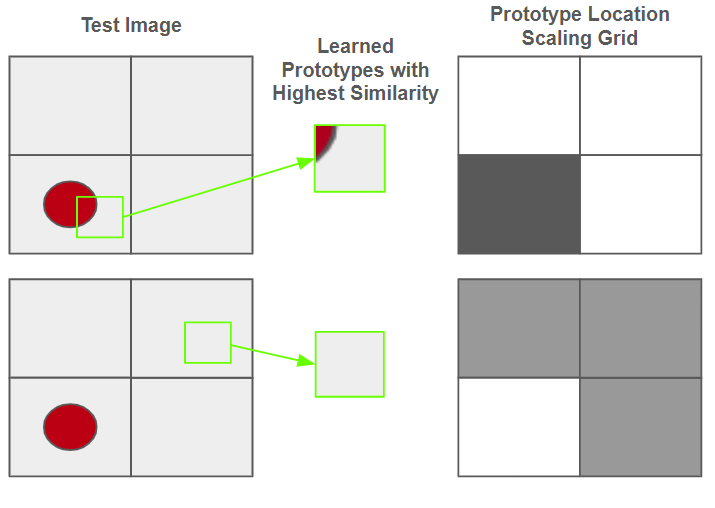}
    \caption{ Notional representation of ProtoLNet's \citep{barnes2022looks} prototype-based reasoning for an example location-relevant learning task classifying which quadrant the object is located. We find those prototypes with the highest similarity to the test image. The location scaling grid shows where each prototype is important for this classification.)}
    \label{fig:mjo}
\end{figure}

\citet{barnes2022looks} extend ProtoPNet \citep{chen2019looks} to generate prototypes along with learned \textit{location scaling} components for the classification of the phases of the Madden Julian Oscillation (MJO), an important atmospheric phenomenon impacting tropical weather at monthly scales, using environmental features, such as wind speed and longwave radiation. In many geoscience applications, the location of features or phenomena is important. This may not hold true for standard image classification applications using a standard ProtoPNet model. For example, in bird classification, the input image can share similarities to a learned prototypical part such as a red beak. However the location of the red beak in the input image does not matter, it is only necessary the input image contain parts similar to a red beak. By contrast, in climate applications in which the inputs are gridded climate data rather than natural images, the absolute location (e.g., near the poles vs.\ near the equator) may be critical for classification. The authors develop ProtoLNet, a method that learns prototype patches along with the location in the image where the prototype is important. For each phase of the MJO, the authors generate prototypical patterns of the environmental features and the associated location where these patterns are important. Figure \ref{fig:mjo} represents a notional diagram of ProtoLNet's reasoning  for a simple location-relevant learning task.  The location scaling grid shows where these prototypes are important, however, an important consideration is that, one location scaling grid is identically applied for all features in the data; in the MJO use-case, prototypical patterns of both wind speed and radiation are required to be present in the same location rather than in unique locations for each feature. 

\section{Discussion} \label{sec:dicussion}
In this section, we first discuss the unique properties of geoscientific data and explore how potential geoscience learning tasks may benefit from the use of the prototype-based XAI methods highlighted above. We also discuss literature that addresses the general limitations of prototype-based explanations and XAI in general, considerations for evaluating their level of interpretability, and different cautions to be aware of when using these methods in the scientific domain. \\

Geoscientific data has unique characteristics that differ from the standard natural image or language data typically used in prototype-based XAI research and thus requires particular attention when using prototype-based techniques. For example, geoscientific data often include both a temporal component, gridded spatial data contain spatial autocorrelations, datasets are often high-dimensional or contain data from multiple spectral channels. Data sets may be multimodal, with information combined from multiple sources or sensing modalities, and both observational and simulation-generated data sets can be massive in volume \citep{karpatne2018machine}. Unlike natural images that usually contain objects of sharp edges, colors or features over a distinct background, geoscientific phenomena tend to be multi-spatial scale and have ambiguous boundaries such as weather fronts at the meso-scale level ($\approx 10^{-1}$ to $10^{2}$ km) and tropical cyclones at the synoptic-scale level ($\approx 10^{3}$ to $10^{4}$ km). Typically, when we treat gridded climate data (for example, a latitude-longitude grid of temperature,  wind speed, and other meteorological variables) like natural images in computer vision, each channel (temperature, wind speed, etc.) is often a separate, independent feature in a particular sample \cite{barnes2022looks}. In contrast, natural RGB images use three channels to represent one color. Thus it may not be appropriate to process or interpret gridded climate data in the same manner as RGB natural images.  In addition, many geoscientific events of scientific interest lack adequate training data \cite{ham2005investigation}; for example in the case natural disasters such as tropical cyclones and flash floods, observational datasets will typically contain very few instances of extreme events \citep{kuglitsch2022artificial, karpatne2018machine}. \\

The applications of image-sized prototypes \citep{li2018deep} can be appropriate for climate and weather learning tasks that require understanding global modes of a target class. Modes derived using Principal Component Analysis (PCA; also referred to as Empirical Orthogonal Function, or EOF, analysis), are commonly used in climate science to analyze climate patterns \cite{hannachi2007empirical, kao2009contrasting} such as El Ni\~{n}o \cite{Wang2017elnino}. Image-sized prototypes offer an alternative method for deriving climate modes, without the constraint of orthogonality imposed in PCA. In contrast, patch prototypes \citep{li2018deep} can be used to investigate localized features or patterns within a larger input grid. For example in climate prediction tasks, an image-sized prototype  can represent a global climate mode (e.g., El Ni\~{n}o phase), while a patch prototype may point to localized features in a specific region (e.g., sea surface temperature patterns in the eastern Pacific) \cite{rivera2023explained}. Deformable prototypes \citep{donnelly2022deformable} enable a flexible configuration of prototype patch clusters within the images. Deformable prototypes would be appropriate for modeling multi-scale interactions such as precipitation events \cite{tan2024deep, prein2023multi} where differently sized and organized prototype patches can inform prototypical patterns for a range of spatial scales. \\

Geoscientific tasks often involve using DL for spatio-temporal forecasting such as short-term weather variables \cite{suleman2022short}, turbulent flow \cite{wang2020towards} and climate oscillations \cite{geng2021spatiotemporal,wang2023interpretable} where generating sequential or temporal explanations can help identify critical events for different forecast lead-times. Originally tested on sequence classification, the  sequential prototype approach of \citet{ming2019interpretable} could be adapted for time-series forecasting. In this approach, a DL model would be able to generate explanations for a forecast depending on the presence of key events, identified by their similarity to salient prototypical sub-sequences. Extending the work in \citet{barnes2022looks} to explore prototypes with individual feature (channel)-specific location scaling grids may help gain more insight into the role specific individual meteorological variables play in climate phenomena. While prototype-based methods have been developed for spatial and temporal data, there has been limited work on combining these approaches for interpretable spatio-temporal data analysis.\\

Prototype-anomaly scores \citep{li2023prototypes} can be useful in detecting anomalous changes in the testing data or whether it is out-of-distribution in an unsupervised manner. Anomaly scores may be useful in the case of climate research, where data is usually non-stationary given natural and anthropogenic climate change (i.e true distribution of the data is always evolving). Learning support prototypes \citep{wang2023learning} can help distinguish harder to discern patterns, which may be useful in detecting climate and weather phenomena that contain more subtle distinguishing characteristics and are harder to predict, such as atmospheric rivers \cite{chapman2019improving} and tropical cyclones \cite{galea2023tcdetect}. Data points with characteristics that are more similar to support prototypes compared to trivial prototypes can point to data points with salient characteristics. For extreme event prediction such as cyclone intensification \cite{xu2021deep} or extreme precipitation events \cite{franch2020precipitation}, one can also consider using the techniques developed in \citet{singh2021these} and \citet{nauta2021neural} to further understand the decision making process when learning tasks may rely on negative feature contributions or benefit from binary tree-like decision making. In standard prototype approaches, models make decisions that are solely based on positive similarity of the input image to a prototype. However in applications where an input image does not contain a crucial prototypical pattern, using negative reasoning or binary tree-like reasoning may benefit the model's prediction accuracy. This can be especially important in extreme event detection and related high-impact decision-making \cite{mcgovern2017using}. \\

In the case of hyperspectral satellite imagery, a common data-type in geoscientific research \cite{rolf2024mission}, techniques are developed to reduce their large dimensionality \cite{santara2017bass}. The motivation behind this is to gain insight into distinguishing features that may not be visible in all spectral bands of a satellite image and find specific salient bands that contain relevant information for a task, in contrast with standard RGB image channels. Generating multi-variable prototype-based explanations as described in \citet{ghosal2021multi} may be a promising alternative method to investigate prototypical patterns within individual spectral channels and the relevant multi-spectral prototypical patterns that are salient for a model's decision process. In general, standard feature importance techniques point to how single-variable features contribute to a model prediction, a common research motivation in geosciences \cite{o2018using,qiu2018feature, wei2023tropical}. Generating multi-variable explanations can provide additional information on how features both individually and in combination contribute to a model prediction, which can be especially important for geoscientific phenomena that tend to be driven by complex multiple environmental features or forcings. \\

While we discuss the potential opportunities for adapting prototype-based XAI techniques in the geosciences, we must also consider evaluating the utility of prototype-based explanations when using them in any scientific domain. For prototype-based explanations, in particular, it is important to avoid generating a large number of prototypes to prevent over-complicating explanations. Though we need the prototypes to be generally representative of the data, prototypes should also be sparse in nature and provide a simple `explanation'.  One can consider including pruning approaches noted in \citet{nauta2021neural} and \citet{chen2019looks} that can generate simpler and non-duplicative prototype based explanations. Domain scientists should play integral roles in evaluating the robustness and reliability of the generated prototype-based explanations for a learning task. For example, including a domain specialist in the human-in-the-loop approach described in \citet{ming2019interpretable} may help reduce the redundancy of the model-generated prototypes to making the explanation easier to comprehend. The case studies discussed above vary in the types of prototypes generated, model and training complexity and the generated final explanation. Depending on the specific learning task and the level of complexity of the explanation that is needed, domain scientists can aid in choosing the appropriate method such that prototypes provide an informative yet concise explanation during the model decision-making process. \\

\citet{hoffmann2021looks} demonstrate the potential pitfalls of using prototype-based networks for generating explanations. The authors find that introducing artifacts into the input data that are imperceptible to humans, can drastically alter the explanations despite the image and the altered image looking virtually identical to each other. This is an essential critique of prototype-based methods because they rely on comparing the similarity of the latent representation of a training sample with the test sample to establish an `explanation'. Artifacts can alter the latent representation significantly which can alter the `explanation' even if a human would mark the altered training sample as similar to the testing sample. To combat these issues, the authors recommend implementing data augmentation techniques and adversarial training to prevent the introduction of artifacts and train the model to recognize the samples augmented with artifacts. Similar pitfalls arise in using traditional post hoc XAI techniques. \citet{ghorbani2019interpretation} and \citet{alvarez2018robustness} show that perturbations to input data produce varied `explanations' when using post hoc XAI techniques indicating a similar risk in their robustness and stability. \citet{kim2022hive} establish a framework, HIVE: Human Interpretability of Visual Explanations that assesses how explanations derived from AI models can help or hinder human decision making. \citet{huang2023evaluation} establish a quantitative interpretability benchmark metric to follow when interrogating the consistency and stability of the prototype-based explanations. Rigorous testing and robustness evaluation of the prototype-based XAI methods should be included in best practices when employing these methods in geosciences, especially when these methods are involved with generating scientific insight and/or involved with high-stakes decision-making.

\section{Conclusion}

In this review, we present case studies from the emerging field of prototype-based XAI, a set of methods that aim to develop built-in interpretability for complex deep learning architectures. These case studies show promising avenues for further development for use-cases in geoscientific research, where there has been significant recent progress through the use of AI. We include case studies that focus on the the development of the prototypes and their visualizations, studies that explore the different form of prototypes, and studies that focus on using prototypes in different types of learning tasks. By adapting these methods and tailoring them for geoscience specific domain tasks, we can generate accurate, reliable, and insightful models. When considering the methods showcased for geosciences applications, one should always consider what the end-goal task is and what form of explanation would be most useful for the specific task. We include potential avenues to use prototype-based methods for geoscientists to explore in their respective research domains. We also include a discussion on studies that have developed rigorous testing and evaluation frameworks to evaluate the robustness and test the integrity of prototype-based explanations when employing these techniques. \\

Though this review focuses on learning tasks with geoscientific data, the points raised in this review also apply more generally to other scientific disciplines that share similar research tasks. Most prototype-based XAI approaches have been developed for standard natural image, text and natural language benchmark datasets. With the increasing adoption of ML in the sciences, there is greater need for XAI tools designed with the unique properties of scientific data and needs of scientific researchers in mind. Such tools, especially when developed in collaboration between scientists and XAI researchers, have the potential to advance both fields.


\section*{Acknowledgements}

This work was supported by the SciAI Center, and funded by the Office of Naval Research (ONR), under Grant Number N00014-23-1-2729. The authors thank the students in EEPS-DATA 1720 (Brown University, Spring 2024) for their peer feedback on early drafts of this review. 


\section*{Impact Statement}

The prototype-based XAI techniques discussed in this work may be implemented with the goal of increasing trustworthiness in scientific applications of ML, especially those that involve high stakes decision-making in societal contexts. Broader Impacts are discussed in the Discussion section of the main text, including our recommendation that generated prototype-based explanations be thoroughly evaluated, with inputs from domain experts, for robustness and reliability before using them as justifications in real-world scenarios. 





\bibliography{bibliography}

\begin{thebibliography}{71}
\providecommand{\natexlab}[1]{#1}
\providecommand{\url}[1]{\texttt{#1}}
\expandafter\ifx\csname urlstyle\endcsname\relax
  \providecommand{\doi}[1]{doi: #1}\else
  \providecommand{\doi}{doi: \begingroup \urlstyle{rm}\Url}\fi

\bibitem[Adebayo et~al.(2018)Adebayo, Gilmer, Muelly, Goodfellow, Hardt, and Kim]{adebayo2018sanity}
Adebayo, J., Gilmer, J., Muelly, M., Goodfellow, I., Hardt, M., and Kim, B.
\newblock Sanity checks for saliency maps.
\newblock \emph{Advances in Neural Information Processing Systems}, 31, 2018.

\bibitem[Alvarez-Melis \& Jaakkola(2018)Alvarez-Melis and Jaakkola]{alvarez2018robustness}
Alvarez-Melis, D. and Jaakkola, T.~S.
\newblock On the robustness of interpretability methods.
\newblock \emph{arXiv preprint arXiv:1806.08049}, 2018.

\bibitem[Bach et~al.(2015)Bach, Binder, Montavon, Klauschen, M{\"u}ller, and Samek]{bach2015pixel}
Bach, S., Binder, A., Montavon, G., Klauschen, F., M{\"u}ller, K.-R., and Samek, W.
\newblock On pixel-wise explanations for non-linear classifier decisions by layer-wise relevance propagation.
\newblock \emph{PloS one}, 10\penalty0 (7):\penalty0 e0130140, 2015.

\bibitem[Barnes et~al.(2022)Barnes, Barnes, Martin, and Rader]{barnes2022looks}
Barnes, E.~A., Barnes, R.~J., Martin, Z.~K., and Rader, J.~K.
\newblock This looks like that there: Interpretable neural networks for image tasks when location matters.
\newblock \emph{Artificial Intelligence for the Earth Systems}, 1\penalty0 (3):\penalty0 e220001, 2022.

\bibitem[Beucler et~al.(2021)Beucler, Pritchard, Rasp, Ott, Baldi, and Gentine]{beucler2021enforcing}
Beucler, T., Pritchard, M., Rasp, S., Ott, J., Baldi, P., and Gentine, P.
\newblock Enforcing analytic constraints in neural networks emulating physical systems.
\newblock \emph{Physical Review Letters}, 126\penalty0 (9):\penalty0 098302, 2021.

\bibitem[Bi et~al.(2023)Bi, Xie, Zhang, Chen, Gu, and Tian]{bi2023accurate}
Bi, K., Xie, L., Zhang, H., Chen, X., Gu, X., and Tian, Q.
\newblock Accurate medium-range global weather forecasting with {3D} neural networks.
\newblock \emph{Nature}, 619\penalty0 (7970):\penalty0 533--538, 2023.

\bibitem[Chapman et~al.(2019)Chapman, Subramanian, Delle~Monache, Xie, and Ralph]{chapman2019improving}
Chapman, W., Subramanian, A., Delle~Monache, L., Xie, S., and Ralph, F.
\newblock Improving atmospheric river forecasts with machine learning.
\newblock \emph{Geophysical Research Letters}, 46\penalty0 (17-18):\penalty0 10627--10635, 2019.

\bibitem[Chen et~al.(2019)Chen, Li, Tao, Barnett, Rudin, and Su]{chen2019looks}
Chen, C., Li, O., Tao, D., Barnett, A., Rudin, C., and Su, J.~K.
\newblock This looks like that: deep learning for interpretable image recognition.
\newblock \emph{Advances in Neural Information Processing systems}, 32, 2019.

\bibitem[de~Burgh-Day \& Leeuwenburg(2023)de~Burgh-Day and Leeuwenburg]{de2023machine}
de~Burgh-Day, C.~O. and Leeuwenburg, T.
\newblock Machine learning for numerical weather and climate modelling: a review.
\newblock \emph{Geoscientific Model Development}, 16\penalty0 (22):\penalty0 6433--6477, 2023.

\bibitem[Donnelly et~al.(2022)Donnelly, Barnett, and Chen]{donnelly2022deformable}
Donnelly, J., Barnett, A.~J., and Chen, C.
\newblock Deformable protopnet: An interpretable image classifier using deformable prototypes.
\newblock In \emph{Proceedings of the IEEE/CVF Conference on Computer Vision and Pattern Recognition}, pp.\  10265--10275, 2022.

\bibitem[Franch et~al.(2020)Franch, Nerini, Pendesini, Coviello, Jurman, and Furlanello]{franch2020precipitation}
Franch, G., Nerini, D., Pendesini, M., Coviello, L., Jurman, G., and Furlanello, C.
\newblock Precipitation nowcasting with orographic enhanced stacked generalization: Improving deep learning predictions on extreme events.
\newblock \emph{Atmosphere}, 11\penalty0 (3):\penalty0 267, 2020.

\bibitem[Galea et~al.(2023)Galea, Kunkel, and Lawrence]{galea2023tcdetect}
Galea, D., Kunkel, J., and Lawrence, B.~N.
\newblock Tcdetect: a new method of detecting the presence of tropical cyclones using deep learning.
\newblock \emph{Artificial Intelligence for the Earth Systems}, 2\penalty0 (3):\penalty0 e220045, 2023.

\bibitem[Geng \& Wang(2021)Geng and Wang]{geng2021spatiotemporal}
Geng, H. and Wang, T.
\newblock Spatiotemporal model based on deep learning for enso forecasts.
\newblock \emph{Atmosphere}, 12\penalty0 (7):\penalty0 810, 2021.

\bibitem[Ghorbani et~al.(2019)Ghorbani, Abid, and Zou]{ghorbani2019interpretation}
Ghorbani, A., Abid, A., and Zou, J.
\newblock Interpretation of neural networks is fragile.
\newblock In \emph{Proceedings of the AAAI conference on Artificial Intelligence}, volume~33, pp.\  3681--3688, 2019.

\bibitem[Ghosal \& Abbasi-Asl(2021)Ghosal and Abbasi-Asl]{ghosal2021multi}
Ghosal, G.~R. and Abbasi-Asl, R.
\newblock Multi-modal prototype learning for interpretable multivariable time series classification.
\newblock \emph{arXiv preprint arXiv:2106.09636}, 2021.

\bibitem[Gilpin et~al.(2018)Gilpin, Bau, Yuan, Bajwa, Specter, and Kagal]{gilpin2018explaining}
Gilpin, L.~H., Bau, D., Yuan, B.~Z., Bajwa, A., Specter, M., and Kagal, L.
\newblock Explaining explanations: An overview of interpretability of machine learning.
\newblock In \emph{2018 IEEE 5th International Conference on data science and advanced analytics (DSAA)}, pp.\  80--89. IEEE, 2018.

\bibitem[Griffin et~al.(2022)Griffin, Wimmers, and Velden]{griffin2022predicting}
Griffin, S.~M., Wimmers, A., and Velden, C.~S.
\newblock Predicting rapid intensification in {North Atlantic} and eastern {North Pacific} tropical cyclones using a convolutional neural network.
\newblock \emph{Weather and Forecasting}, 37\penalty0 (8):\penalty0 1333--1355, 2022.

\bibitem[Ham et~al.(2005)Ham, Chen, Crawford, and Ghosh]{ham2005investigation}
Ham, J., Chen, Y., Crawford, M., and Ghosh, J.
\newblock Investigation of the random forest framework for classification of hyperspectral data.
\newblock \emph{IEEE Transactions on Geoscience and Remote Sensing}, 43\penalty0 (3):\penalty0 492--501, 2005.

\bibitem[Ham et~al.(2019)Ham, Kim, and Luo]{ham2019deep}
Ham, Y.-G., Kim, J.-H., and Luo, J.-J.
\newblock Deep learning for multi-year {ENSO} forecasts.
\newblock \emph{Nature}, 573\penalty0 (7775):\penalty0 568--572, 2019.

\bibitem[Hannachi et~al.(2007)Hannachi, Jolliffe, and Stephenson]{hannachi2007empirical}
Hannachi, A., Jolliffe, I.~T., and Stephenson, D.~B.
\newblock Empirical orthogonal functions and related techniques in atmospheric science: A review.
\newblock \emph{International Journal of Climatology: A Journal of the Royal Meteorological Society}, 27\penalty0 (9):\penalty0 1119--1152, 2007.

\bibitem[Hilburn et~al.(2020)Hilburn, Ebert-Uphoff, and Miller]{hilburn2020development}
Hilburn, K.~A., Ebert-Uphoff, I., and Miller, S.~D.
\newblock Development and interpretation of a neural-network-based synthetic radar reflectivity estimator using goes-r satellite observations.
\newblock \emph{Journal of Applied Meteorology and Climatology}, 60\penalty0 (1):\penalty0 3--21, 2020.

\bibitem[Hoffmann et~al.(2021)Hoffmann, Fanconi, Rade, and Kohler]{hoffmann2021looks}
Hoffmann, A., Fanconi, C., Rade, R., and Kohler, J.
\newblock This looks like that... does it? shortcomings of latent space prototype interpretability in deep networks.
\newblock \emph{arXiv preprint arXiv:2105.02968}, 2021.

\bibitem[Huang et~al.(2023)Huang, Xue, Huang, Zhang, Song, Jing, and Song]{huang2023evaluation}
Huang, Q., Xue, M., Huang, W., Zhang, H., Song, J., Jing, Y., and Song, M.
\newblock Evaluation and improvement of interpretability for self-explainable part-prototype networks.
\newblock In \emph{Proceedings of the IEEE/CVF International Conference on Computer Vision}, pp.\  2011--2020, 2023.

\bibitem[Iglesias-Suarez et~al.(2024)Iglesias-Suarez, Gentine, Solino-Fernandez, Beucler, Pritchard, Runge, and Eyring]{iglesias2024causally}
Iglesias-Suarez, F., Gentine, P., Solino-Fernandez, B., Beucler, T., Pritchard, M., Runge, J., and Eyring, V.
\newblock Causally-informed deep learning to improve climate models and projections.
\newblock \emph{Journal of Geophysical Research: Atmospheres}, 129\penalty0 (4):\penalty0 e2023JD039202, 2024.

\bibitem[Jakubik et~al.(2023)Jakubik, Roy, Phillips, Fraccaro, Godwin, Zadrozny, Szwarcman, Gomes, Nyirjesy, Edwards, et~al.]{jakubik2023foundation}
Jakubik, J., Roy, S., Phillips, C., Fraccaro, P., Godwin, D., Zadrozny, B., Szwarcman, D., Gomes, C., Nyirjesy, G., Edwards, B., et~al.
\newblock Foundation models for generalist geospatial artificial intelligence.
\newblock \emph{arXiv preprint arXiv:2310.18660}, 2023.

\bibitem[Kao \& Yu(2009)Kao and Yu]{kao2009contrasting}
Kao, H.-Y. and Yu, J.-Y.
\newblock Contrasting eastern-pacific and central-pacific types of enso.
\newblock \emph{Journal of Climate}, 22\penalty0 (3):\penalty0 615--632, 2009.

\bibitem[Karpatne et~al.(2019)Karpatne, Ebert-Uphoff, Ravela, Babaie, and Kumar]{karpatne2018machine}
Karpatne, A., Ebert-Uphoff, I., Ravela, S., Babaie, H.~A., and Kumar, V.
\newblock Machine learning for the geosciences: Challenges and opportunities.
\newblock \emph{IEEE Transactions on Knowledge and Data Engineering}, 31\penalty0 (8):\penalty0 1544--1554, 2019.

\bibitem[Keane \& Kenny(2019)Keane and Kenny]{keane2019case}
Keane, M.~T. and Kenny, E.~M.
\newblock How case-based reasoning explains neural networks: A theoretical analysis of {XAI} using post-hoc explanation-by-example from a survey of {ANN-CBR} twin-systems.
\newblock In \emph{Case-Based Reasoning Research and Development: 27th International Conference, ICCBR 2019, Otzenhausen, Germany, September 8--12, 2019, Proceedings 27}, pp.\  155--171. Springer, 2019.

\bibitem[Kim et~al.(2024)Kim, Ham, Kim, Li, and Ma]{kim2024improvement}
Kim, J.-H., Ham, Y.-G., Kim, D., Li, T., and Ma, C.
\newblock Improvement in forecasting short-term tropical cyclone intensity change and their rapid intensification using deep learning.
\newblock \emph{Artificial Intelligence for the Earth Systems}, 3\penalty0 (2):\penalty0 e230052, 2024.

\bibitem[Kim et~al.(2022)Kim, Meister, Ramaswamy, Fong, and Russakovsky]{kim2022hive}
Kim, S.~S., Meister, N., Ramaswamy, V.~V., Fong, R., and Russakovsky, O.
\newblock {HIVE}: Evaluating the human interpretability of visual explanations.
\newblock In \emph{European Conference on Computer Vision}, pp.\  280--298. Springer, 2022.

\bibitem[Kuglitsch et~al.(2022)Kuglitsch, Albayrak, Aquino, Craddock, Gill, Kanwar, Koul, Ma, Marti, Menon, et~al.]{kuglitsch2022artificial}
Kuglitsch, M., Albayrak, A., Aquino, R., Craddock, A., Gill, J.~E., Kanwar, R., Koul, A., Ma, J., Marti, A., Menon, M., et~al.
\newblock Artificial intelligence for disaster risk reduction: Opportunities, challenges, and prospects.
\newblock \emph{World Meteorological Organization (WMO) Bulletin}, 71\penalty0 (1):\penalty0 30--37, 2022.

\bibitem[Lam et~al.(2023)Lam, Sanchez-Gonzalez, Willson, Wirnsberger, Fortunato, Alet, Ravuri, Ewalds, Eaton-Rosen, Hu, et~al.]{lam2023learning}
Lam, R., Sanchez-Gonzalez, A., Willson, M., Wirnsberger, P., Fortunato, M., Alet, F., Ravuri, S., Ewalds, T., Eaton-Rosen, Z., Hu, W., et~al.
\newblock Learning skillful medium-range global weather forecasting.
\newblock \emph{Science}, 382\penalty0 (6677):\penalty0 1416--1421, 2023.

\bibitem[Li et~al.(2023)Li, Jentsch, and M{\"u}ller]{li2023prototypes}
Li, B., Jentsch, C., and M{\"u}ller, E.
\newblock Prototypes as explanation for time series anomaly detection.
\newblock \emph{arXiv preprint arXiv:2307.01601}, 2023.

\bibitem[Li et~al.(2018)Li, Liu, Chen, and Rudin]{li2018deep}
Li, O., Liu, H., Chen, C., and Rudin, C.
\newblock Deep learning for case-based reasoning through prototypes: A neural network that explains its predictions.
\newblock In \emph{Proceedings of the AAAI Conference on Artificial Intelligence}, volume~32, 2018.

\bibitem[Lipton(2018)]{lipton2018mythos}
Lipton, Z.~C.
\newblock The mythos of model interpretability: In machine learning, the concept of interpretability is both important and slippery.
\newblock \emph{Queue}, 16\penalty0 (3):\penalty0 31--57, 2018.

\bibitem[Lundberg \& Lee(2017)Lundberg and Lee]{lundberg2017unified}
Lundberg, S.~M. and Lee, S.-I.
\newblock A unified approach to interpreting model predictions.
\newblock \emph{Advances in Neural Information Processing Systems}, 30, 2017.

\bibitem[Mamalakis et~al.(2020)Mamalakis, Ebert-Uphoff, and Barnes]{mamalakis2020explainable}
Mamalakis, A., Ebert-Uphoff, I., and Barnes, E.~A.
\newblock Explainable artificial intelligence in meteorology and climate science: Model fine-tuning, calibrating trust and learning new science.
\newblock In \emph{International Workshop on Extending Explainable AI Beyond Deep Models and Classifiers}, pp.\  315--339. Springer, 2020.

\bibitem[Mamalakis et~al.(2022)Mamalakis, Barnes, and Ebert-Uphoff]{mamalakis2022investigating}
Mamalakis, A., Barnes, E.~A., and Ebert-Uphoff, I.
\newblock Investigating the fidelity of explainable artificial intelligence methods for applications of convolutional neural networks in geoscience.
\newblock \emph{Artificial Intelligence for the Earth Systems}, 1\penalty0 (4):\penalty0 e220012, 2022.

\bibitem[Mayer \& Barnes(2021)Mayer and Barnes]{mayer2021subseasonal}
Mayer, K.~J. and Barnes, E.~A.
\newblock Subseasonal forecasts of opportunity identified by an explainable neural network.
\newblock \emph{Geophysical Research Letters}, 48\penalty0 (10):\penalty0 e2020GL092092, 2021.

\bibitem[McGovern et~al.(2017)McGovern, Elmore, Gagne, Haupt, Karstens, Lagerquist, Smith, and Williams]{mcgovern2017using}
McGovern, A., Elmore, K.~L., Gagne, D.~J., Haupt, S.~E., Karstens, C.~D., Lagerquist, R., Smith, T., and Williams, J.~K.
\newblock Using artificial intelligence to improve real-time decision-making for high-impact weather.
\newblock \emph{Bulletin of the American Meteorological Society}, 98\penalty0 (10):\penalty0 2073--2090, 2017.

\bibitem[McGovern et~al.(2019)McGovern, Lagerquist, Gagne, Jergensen, Elmore, Homeyer, and Smith]{mcgovern2019making}
McGovern, A., Lagerquist, R., Gagne, D.~J., Jergensen, G.~E., Elmore, K.~L., Homeyer, C.~R., and Smith, T.
\newblock Making the black box more transparent: Understanding the physical implications of machine learning.
\newblock \emph{Bulletin of the American Meteorological Society}, 100\penalty0 (11):\penalty0 2175--2199, 2019.

\bibitem[Ming et~al.(2019)Ming, Xu, Qu, and Ren]{ming2019interpretable}
Ming, Y., Xu, P., Qu, H., and Ren, L.
\newblock Interpretable and steerable sequence learning via prototypes.
\newblock In \emph{Proceedings of the 25th ACM SIGKDD International Conference on Knowledge Discovery \& Data Mining}, pp.\  903--913, 2019.

\bibitem[Molina et~al.(2023)Molina, O’Brien, Anderson, Ashfaq, Bennett, Collins, Dagon, Restrepo, and Ullrich]{molina2023review}
Molina, M.~J., O’Brien, T.~A., Anderson, G., Ashfaq, M., Bennett, K.~E., Collins, W.~D., Dagon, K., Restrepo, J.~M., and Ullrich, P.~A.
\newblock A review of recent and emerging machine learning applications for climate variability and weather phenomena.
\newblock \emph{Artificial Intelligence for the Earth Systems}, 2\penalty0 (4):\penalty0 220086, 2023.

\bibitem[Nauta et~al.(2021)Nauta, Van~Bree, and Seifert]{nauta2021neural}
Nauta, M., Van~Bree, R., and Seifert, C.
\newblock Neural prototype trees for interpretable fine-grained image recognition.
\newblock In \emph{Proceedings of the IEEE/CVF Conference on Computer Vision and Pattern Recognition}, pp.\  14933--14943, 2021.

\bibitem[Neely et~al.(2021)Neely, Schouten, Bleeker, and Lucic]{neely2021order}
Neely, M., Schouten, S.~F., Bleeker, M.~J., and Lucic, A.
\newblock Order in the court: Explainable {AI} methods prone to disagreement.
\newblock \emph{arXiv preprint arXiv:2105.03287}, 2021.

\bibitem[Nguyen et~al.(2023)Nguyen, Brandstetter, Kapoor, Gupta, and Grover]{nguyen2023climax}
Nguyen, T., Brandstetter, J., Kapoor, A., Gupta, J.~K., and Grover, A.
\newblock {C}lima{X}: A foundation model for weather and climate.
\newblock In \emph{Proceedings of the 40th International Conference on Machine Learning (ICML)}, volume 202, pp.\  25904--25938, 23--29 Jul 2023.

\bibitem[O'Gorman \& Dwyer(2018)O'Gorman and Dwyer]{o2018using}
O'Gorman, P.~A. and Dwyer, J.~G.
\newblock Using machine learning to parameterize moist convection: Potential for modeling of climate, climate change, and extreme events.
\newblock \emph{Journal of Advances in Modeling Earth Systems}, 10\penalty0 (10):\penalty0 2548--2563, 2018.

\bibitem[Pathak et~al.(2022)Pathak, Subramanian, Harrington, Raja, Chattopadhyay, Mardani, Kurth, Hall, Li, Azizzadenesheli, et~al.]{pathak2022fourcastnet}
Pathak, J., Subramanian, S., Harrington, P., Raja, S., Chattopadhyay, A., Mardani, M., Kurth, T., Hall, D., Li, Z., Azizzadenesheli, K., et~al.
\newblock {FourCastNet}: A global data-driven high-resolution weather model using adaptive {Fourier} neural operators.
\newblock \emph{arXiv preprint arXiv:2202.11214}, 2022.

\bibitem[Prabhat et~al.(2020)Prabhat, Kashinath, Mudigonda, Kim, Kapp-Schwoerer, Graubner, Karaismailoglu, von Kleist, Kurth, Greiner, et~al.]{prabhat2020climatenet}
Prabhat, Kashinath, K., Mudigonda, M., Kim, S., Kapp-Schwoerer, L., Graubner, A., Karaismailoglu, E., von Kleist, L., Kurth, T., Greiner, A., et~al.
\newblock Climatenet: An expert-labelled open dataset and deep learning architecture for enabling high-precision analyses of extreme weather.
\newblock \emph{Geoscientific Model Development Discussions}, 2020:\penalty0 1--28, 2020.

\bibitem[Prein et~al.(2023)Prein, Mooney, and Done]{prein2023multi}
Prein, A.~F., Mooney, P.~A., and Done, J.~M.
\newblock The multi-scale interactions of atmospheric phenomenon in mean and extreme precipitation.
\newblock \emph{Earth's Future}, 11\penalty0 (11):\penalty0 e2023EF003534, 2023.

\bibitem[Qiu et~al.(2018)Qiu, Schmitt, Mou, Ghamisi, and Zhu]{qiu2018feature}
Qiu, C., Schmitt, M., Mou, L., Ghamisi, P., and Zhu, X.~X.
\newblock Feature importance analysis for local climate zone classification using a residual convolutional neural network with multi-source datasets.
\newblock \emph{Remote Sensing}, 10\penalty0 (10):\penalty0 1572, 2018.

\bibitem[Reichstein et~al.(2019)Reichstein, Camps-Valls, Stevens, Jung, Denzler, Carvalhais, and Prabhat]{reichstein2019deep}
Reichstein, M., Camps-Valls, G., Stevens, B., Jung, M., Denzler, J., Carvalhais, N., and Prabhat, f.
\newblock Deep learning and process understanding for data-driven {Earth} system science.
\newblock \emph{Nature}, 566\penalty0 (7743):\penalty0 195--204, 2019.

\bibitem[Ribeiro et~al.(2016)Ribeiro, Singh, and Guestrin]{ribeiro2016should}
Ribeiro, M.~T., Singh, S., and Guestrin, C.
\newblock "why should i trust you?" explaining the predictions of any classifier.
\newblock In \emph{Proceedings of the 22nd ACM SIGKDD International Conference on Knowledge Discovery and Data Mining}, pp.\  1135--1144, 2016.

\bibitem[Rivera~Tello et~al.(2023)Rivera~Tello, Takahashi, and Karamperidou]{rivera2023explained}
Rivera~Tello, G.~A., Takahashi, K., and Karamperidou, C.
\newblock Explained predictions of strong eastern pacific el ni{\~n}o events using deep learning.
\newblock \emph{Scientific Reports}, 13\penalty0 (1):\penalty0 21150, 2023.

\bibitem[Rolf et~al.(2024)Rolf, Klemmer, Robinson, and Kerner]{rolf2024mission}
Rolf, E., Klemmer, K., Robinson, C., and Kerner, H.
\newblock Mission critical--satellite data is a distinct modality in machine learning.
\newblock \emph{arXiv preprint arXiv:2402.01444}, 2024.

\bibitem[Rolnick et~al.(2022)Rolnick, Donti, Kaack, Kochanski, Lacoste, Sankaran, Ross, Milojevic-Dupont, Jaques, Waldman-Brown, et~al.]{rolnick2022tackling}
Rolnick, D., Donti, P.~L., Kaack, L.~H., Kochanski, K., Lacoste, A., Sankaran, K., Ross, A.~S., Milojevic-Dupont, N., Jaques, N., Waldman-Brown, A., et~al.
\newblock Tackling climate change with machine learning.
\newblock \emph{ACM Computing Surveys (CSUR)}, 55\penalty0 (2):\penalty0 1--96, 2022.

\bibitem[Rudin(2019)]{rudin2019stop}
Rudin, C.
\newblock Stop explaining black box machine learning models for high stakes decisions and use interpretable models instead.
\newblock \emph{Nature Machine Intelligence}, 1\penalty0 (5):\penalty0 206--215, 2019.

\bibitem[Santara et~al.(2017)Santara, Mani, Hatwar, Singh, Garg, Padia, and Mitra]{santara2017bass}
Santara, A., Mani, K., Hatwar, P., Singh, A., Garg, A., Padia, K., and Mitra, P.
\newblock {BASS} net: Band-adaptive spectral-spatial feature learning neural network for hyperspectral image classification.
\newblock \emph{IEEE Transactions on Geoscience and Remote Sensing}, 55\penalty0 (9):\penalty0 5293--5301, 2017.

\bibitem[Singh \& Yow(2021)Singh and Yow]{singh2021these}
Singh, G. and Yow, K.-C.
\newblock These do not look like those: An interpretable deep learning model for image recognition.
\newblock \emph{IEEE Access}, 9:\penalty0 41482--41493, 2021.

\bibitem[Suleman \& Shridevi(2022)Suleman and Shridevi]{suleman2022short}
Suleman, M. A.~R. and Shridevi, S.
\newblock Short-term weather forecasting using spatial feature attention based lstm model.
\newblock \emph{IEEE Access}, 10:\penalty0 82456--82468, 2022.

\bibitem[Tan et~al.(2024)Tan, Huang, and Chen]{tan2024deep}
Tan, J., Huang, Q., and Chen, S.
\newblock Deep learning model based on multi-scale feature fusion for precipitation nowcasting.
\newblock \emph{Geoscientific Model Development}, 17\penalty0 (1):\penalty0 53--69, 2024.

\bibitem[Toms et~al.(2020)Toms, Barnes, and Ebert-Uphoff]{toms2020physically}
Toms, B.~A., Barnes, E.~A., and Ebert-Uphoff, I.
\newblock Physically interpretable neural networks for the geosciences: Applications to earth system variability.
\newblock \emph{Journal of Advances in Modeling Earth Systems}, 12\penalty0 (9):\penalty0 e2019MS002002, 2020.

\bibitem[Wang et~al.(2017)Wang, Deser, Yu, DiNezio, and Clement]{Wang2017elnino}
Wang, C., Deser, C., Yu, J.-Y., DiNezio, P., and Clement, A.
\newblock \emph{El Ni{\~{n}}o and Southern Oscillation (ENSO): A Review}, pp.\  85--106.
\newblock Springer Netherlands, Dordrecht, 2017.

\bibitem[Wang et~al.(2023{\natexlab{a}})Wang, Liu, Chen, Liu, Tian, McCarthy, Frazer, and Carneiro]{wang2023learning}
Wang, C., Liu, Y., Chen, Y., Liu, F., Tian, Y., McCarthy, D., Frazer, H., and Carneiro, G.
\newblock Learning support and trivial prototypes for interpretable image classification.
\newblock In \emph{Proceedings of the IEEE/CVF International Conference on Computer Vision}, pp.\  2062--2072, 2023{\natexlab{a}}.

\bibitem[Wang et~al.(2023{\natexlab{b}})Wang, Fu, Du, Gao, Huang, Liu, Chandak, Liu, Van~Katwyk, Deac, et~al.]{wang2023scientific}
Wang, H., Fu, T., Du, Y., Gao, W., Huang, K., Liu, Z., Chandak, P., Liu, S., Van~Katwyk, P., Deac, A., et~al.
\newblock Scientific discovery in the age of artificial intelligence.
\newblock \emph{Nature}, 620\penalty0 (7972):\penalty0 47--60, 2023{\natexlab{b}}.

\bibitem[Wang et~al.(2023{\natexlab{c}})Wang, Hu, and Li]{wang2023interpretable}
Wang, H., Hu, S., and Li, X.
\newblock An interpretable deep learning enso forecasting model.
\newblock \emph{Ocean-Land-Atmosphere Research}, 2:\penalty0 0012, 2023{\natexlab{c}}.

\bibitem[Wang et~al.(2020)Wang, Kashinath, Mustafa, Albert, and Yu]{wang2020towards}
Wang, R., Kashinath, K., Mustafa, M., Albert, A., and Yu, R.
\newblock Towards physics-informed deep learning for turbulent flow prediction.
\newblock In \emph{Proceedings of the 26th ACM SIGKDD international conference on knowledge discovery \& data mining}, pp.\  1457--1466, 2020.

\bibitem[Wei et~al.(2023)Wei, Fang, and Ge]{wei2023tropical}
Wei, M., Fang, G., and Ge, Y.
\newblock Tropical cyclone genesis prediction based on support vector machine considering effects of multiple meteorological parameters.
\newblock \emph{Journal of Wind Engineering and Industrial Aerodynamics}, 243:\penalty0 105591, 2023.

\bibitem[Xu et~al.(2021)Xu, Balaguru, August, Lalo, Hodas, DeMaria, and Judi]{xu2021deep}
Xu, W., Balaguru, K., August, A., Lalo, N., Hodas, N., DeMaria, M., and Judi, D.
\newblock Deep learning experiments for tropical cyclone intensity forecasts.
\newblock \emph{Weather and Forecasting}, 36\penalty0 (4):\penalty0 1453--1470, 2021.

\bibitem[Yang et~al.(2024)Yang, Hu, Li, Mu, Yu, Xia, Li, Dasgupta, and Xiong]{yang2024interpretable}
Yang, R., Hu, J., Li, Z., Mu, J., Yu, T., Xia, J., Li, X., Dasgupta, A., and Xiong, H.
\newblock Interpretable machine learning for weather and climate prediction: A survey.
\newblock \emph{arXiv preprint arXiv:2403.18864}, 2024.

\bibitem[Zhou et~al.(2021)Zhou, Gandomi, Chen, and Holzinger]{zhou2021evaluating}
Zhou, J., Gandomi, A.~H., Chen, F., and Holzinger, A.
\newblock Evaluating the quality of machine learning explanations: A survey on methods and metrics.
\newblock \emph{Electronics}, 10\penalty0 (5):\penalty0 593, 2021.

\end{thebibliography}
\bibliographystyle{icml2024}

\newpage
\appendix
\onecolumn
\section{Appendix}

\renewcommand{\arraystretch}{2}
\begin{table}[htb!]
        \caption{Column 1 depicts  characteristics and properties of geoscientific data. Column 2 states the specific ML tasks in which they're commonly used. Column 3 states the challenges in using them in standard ML and XAI methods. Column 4 includes an example from the literature for the geoscientific task. For a more in-depth review on the development and usage of machine learning and post-hoc XAI analysis in the geosciences, one can refer to \citet{molina2023review}, \citet{yang2024interpretable},  \citet{de2023machine} and \citet{mamalakis2020explainable}. }
    \label{tab:Summary}
    \centering
    \begin{tabular}{|p{0.2\textwidth} | p{0.15\textwidth} |p{0.25\textwidth} |p{0.25\textwidth} |}
        \hline
        \textbf{Geoscientific Data Characteristics} & \textbf{Geoscientific ML Task} & \textbf{Challenges with Standard ML and XAI techniques} & \textbf{Literature Example}\\
        \hline\hline
         Phenomena of interest have ambiguous boundaries and can blend with image background & geoscientific feature detection & Image recognition models rely on features like defined boundaries or sharp edges over a distinct background & tropical cyclone and atmospheric river segmentation \cite{prabhat2020climatenet} \\
        \hline
         Gridded data contains separate, independent environmental features & climate prediction & Often treated as RGB channels in an image representing a singular color & prediction of strong  El Ni\~{n}o events \cite{rivera2023explained} \\
        \hline
        Hyper- or multi-spectral satellite data & geoscientific feature/channel selection, dimensionality-reduction & High dimensionality with relevant information spread across multiple channels in contrast with RGB natural images & estimating radar reflectivity using remote-sensing \cite{hilburn2020development}\\ 
        \hline
        Spatio-temporal, location-specific patterns, temporal patterns & climate mode variability, spatio-temporal forecasting & Sequential methods are developed to work with text or natural language data (tokens vs. continuous signals) &  sub-seasonal tropical-extratropical circulation forecasting \cite{mayer2021subseasonal}\\
        \hline
        Domain-shift, anomalous data & climate prediction & Poor performance on out of distribution data & climate projections under different climate regimes or distributions \cite{iglesias2024causally} \\
        \hline
        Limited observed samples of a rare or extreme event of interest & extreme weather, climate prediction & Class imbalanced data, poor prediction performance & prediction of extreme rapid-intensification cyclone events \cite{kim2024improvement}\\
        \hline   
    \end{tabular}
\end{table}

\renewcommand{\arraystretch}{1.5}
\begin{table}[htb!]
        \caption{For each case study in Column 1, Column 2 states the explanation each study provides using prototypes and Column 3 states potential geoscientific learning tasks for this method. }
    \label{tab:Summary}
    \centering
    \begin{tabular}{|p{0.2\textwidth} | p{0.4\textwidth} |p{0.3\textwidth} |}
        \hline
        \textbf{Case Study} & \textbf{Summary Explanation} & \textbf{Geoscientific Learning Tasks} \\
        \hline\hline
         \citet{li2018deep} & This input image is overall resembles a learned prototypical image of the target class.  &  climate phase prediction, dimensionality reduction \\
        \hline
         \citet{chen2019looks} & This specific patch in input image resembles a learned prototypical local patch or pattern directly from a patch in a training image. &  generating local feature patterns  \\
        \hline
        \citet{donnelly2022deformable} & This organized cluster of prototypical patches resembles a set of prototypical patches within an image of the target class. & organization of feature patterns, detecting multi-scale feature patterns  \\
        \hline
        \citet{wang2023learning} & This input may resemble prototypes representative of a target class or the target class boundary.  & hard-to-learn feature detection, extreme event detection \\
        \hline
        \citet{ming2019interpretable} & A window of this input sequence resembles a prototypical window sequence in the target class. & temporal forecasting, climate oscillations prediction\\
        \hline
        \citet{ghosal2021multi} & This multi-variable input shares similarities with single variables prototypes and their relationships with each other. & multi-forcing classification, multi-variable feature attribution, multi-spectral imagery classification\\
        \hline
        \citet{li2023prototypes} & This input sequence of data significantly differs from prototypical window sequences in the training data. & anomaly detection, extreme event detection  \\
        \hline
        \citet{nauta2021neural} & The prediction is derived from a tree-like reasoning process on whether the input contains similarities to certain prototypes. & multi-variable prediction tasks \\
        \hline   
        \citet{singh2021these}& This input image contains similar prototypical parts of the target class and does not contain prototypical parts from another class. & climate prediction and classification, extreme event prediction \\
        \hline
        \citet{barnes2022looks} & This input image contains similar prototypical parts of the target class only in specific regions of the image. & generating spatially relevant feature patterns\\
        \hline
    \end{tabular}
\end{table}




\end{document}